%% file: acl2020.tex
\documentclass[11pt,a4paper]{article}
\usepackage[hyperref]{acl2020}
\usepackage{times}
\usepackage{latexsym}
\usepackage{color,soul}
\usepackage{amsmath}
\usepackage{algorithm}
\usepackage{graphicx}
\usepackage[noend]{algpseudocode}
\usepackage{comment}
\usepackage{url}

\aclfinalcopy 

\usepackage{flushend}

\usepackage[draft]{todonotes}   

\title{Multilingual Machine Translation: Closing the Gap between Shared and Language-specific Encoder-Decoders}

\author{Carlos Escolano, Marta R. Costa-juss\`a, Jos\'e A. R. Fonollosa, Mikel Artetxe$\dag$\\
  TALP Research Center, Universitat Polit\`ecnica de Catalunya, Barcelona \\
  $\dag$ IXA NLP Group, University of the Basque Country (UPV/EHU)\\  
  {\tt \{carlos.escolano,marta.ruiz,jose.fonollosa\}@upc.edu}\\{\tt mikel.artetxe@ehu.eus} \\}

\date{}

\begin{document}

\maketitle

\begin{abstract}
State-of-the-art multilingual machine translation relies on a universal encoder-decoder, which requires retraining the entire system to add new languages. In this paper, we propose an alternative approach that is based on language-specific encoder-decoders, and can thus be more easily extended to new languages by learning their corresponding modules. So as to encourage a common interlingua representation, we simultaneously train the $N$ initial languages. Our experiments show that the proposed approach outperforms the universal encoder-decoder by 3.28 BLEU points on average, and when adding new languages, without the need to retrain the rest of the modules. All in all, our work closes the gap between shared and language-specific encoder-decoders, advancing toward modular multilingual machine translation systems that can be flexibly extended in lifelong learning settings.

\end{abstract}

\section{Introduction} \label{sec:intro}
\input{introduction.tex}

\section{Related Work} \label{sec:related}
\input{related-work.tex}



\section{Proposed Method} 
\label{sec:proposed}
\input{multilingual.tex}


\section{Experiments in Multilingual Machine Translation} \label{sec:mt}

\input{mtexperiments.tex}

\section{Conclusions} \label{sec:conclusions}

\input{conclusions.tex}

\section*{Acknowledgments}
This work is supported in part by the Google Faculty Research Award 2018, the Spanish Ministerio de Econom\'ia y Competitividad, the European Regional Development Fund through the  postdoctoral senior grant Ram\'on y Cajal and by the Agencia  Estatal  de  Investigaci\'on through the project EUR2019-103819.

\bibliography{emnlp-ijcnlp-2019}
\bibliographystyle{acl_natbib}



\end{document}

%% file: introduction.tex
Multilingual machine translation is the ability to generate translations automatically across a (large) number of languages. Research in this area has attracted a lot of attention in recent times both from the scientific and industrial community. Under the neural machine translation paradigm \cite{bahdanau2014neural}, the opportunities for improving this area have dramatically expanded. Thanks to the encoder-decoder architecture, there are viable alternatives to expensive pairwise translation based on classic paradigms\footnote{http://www.euromatrixplus.net}. 

The main proposal in this direction is the universal encoder-decoder \cite{johnson2017google} with massive multilingual enhancements \cite{DBLP:journals/corr/abs-1907-05019}. While this approach enables zero-shot translation and is beneficial for low-resource languages, it has multiple drawbacks: (i) the entire system has to be retrained when adding new languages or data; (ii) the quality of translation drops when adding too many languages or for those with the most resources \cite{DBLP:journals/corr/abs-1907-05019}; and (iii) the shared vocabulary grows dramatically when adding a large number of languages (especially when they do not share alphabets). 
Other limitations include the incompatibility of adding multiple modalities such as image or speech. 



In this paper, we propose a new framework that can be incrementally extended to new languages without the aforementioned limitations (\S\ref{sec:proposed}). Our proposal is based on language-specific encoders and decoders that rely on a common intermediate representation space. For that purpose, we simultaneously train the initial \textit{N} languages in all translation directions.  New languages are naturally added to the system by training a new module coupled with any of the existing ones, while new data can be easily added by retraining only the module for the corresponding language.

We evaluate our proposal on three experimental configurations: translation for the initial languages,  translation when adding a new language, and zero-shot translation (\S\ref{sec:mt}). Our results show that the proposed method is better in the first configurations by improving the universal system: by 0.40 BLEU points on average in the initial training and by 3.28 BLEU points on average when adding new languages. However, our proposed system is still lagging behind universal encoder-decoder in zero-shot translation. 


%% file: related-work.tex

Multilingual neural machine translation can refer to translating from one-to-many languages \cite{dong-etal-2015-multi}, from many-to-one \cite{zoph-knight-2016-multi} and many-to-many \cite{johnson2017google}. 
Within the many-to-many paradigm, existing approaches can be further divided into shared or language-specific encoder-decoders, which is the approach that we follow. 

\paragraph{Shared Encoder-Decoder.} \citet{johnson2017google} 
feed a single encoder and decoder with multiple input and output languages. Given a set of languages, a shared architecture has a universal encoder and a universal decoder that are trained on all initial language pairs at once. The model shares parameters, vocabulary and tokenization among languages to ensure that no additional ambiguity is introduced in the representation. 
 This architecture provides a simple framework to develop multilingual systems because it does not require modifications of a standard neural machine translation model, and information is easily shared among the different languages through common parameters. Despite the model's advantages in transfer learning, the use of a shared vocabulary and embedding representation forces the model to employ a vocabulary that includes tokens from all the alphabets used. 
Additionally, recent work \cite{DBLP:journals/corr/abs-1907-05019}, that imposes representational invariance across language, shows increasing the number of languages varies the quality of the languages already in the system (generally enhancing low-resource pairs but being detrimental for high-resource pairs). Some other disadvantages are that the number of parameters related to vocabulary grow with the number of languages with different alphabets and the entire system has to be retrained when adding new languages. 

\paragraph{Language-specific Encoder-Decoders.} 
Approaches within this category may or may not share parameters at some point. 

\textit{Sharing parameters.} \citet{firat2016zero} proposed extending the bilingual recurrent neural machine translation architecture \cite{bahdanau2014neural} to the multilingual case \cite{vazquez-etal-2019-multilingual,lu2018neural} by designing a shared attention-based mechanism between the language-specific encoders and decoders to create a language independent representation.   These architectures provide the flexibility for each language to be trained with its own vocabulary, preventing problems related to the addition of several alphabets in the same model, especially when some of them are underrepresented. However, as the language specific components rely on the shared modules, modifying those components to add a new language or add further data to the system would require retraining the whole system. 
\cite{DBLP:journals/corr/abs-1811-01137} proposes a model based on the addition of new languages to an already trained system by vocabulary adaptation and transfer learning. While limited, it requires some retraining to adapt the model to the new task; this results in the translation quality of existing languages varying when adding new ones.

\textit{No sharing.}  The system proposed by \citet{escolano-etal-2019-bilingual} is trained on language-specific encoders and decoders based on joint training without parameter or vocabulary-sharing and on enforcing a compatible representation between the jointly trained languages. The advantage of the approach is that it does not require retraining to add new languages and increasing the number of languages does not vary the quality of the languages already in the system. However, the system has to be trained on a multi-parallel
corpus and the system does not scale well when there is a large number of languages in the initial system, since all encoders and decoders have to be trained simultaneously.



%% file: multilingual.tex
Our proposed approach trains
a separate encoder and decoder for each of the $N$ languages available without requiring multi-parallel corpus. We do not share any parameter across these modules, which allows to add new languages incrementally without retraining the entire system.

\subsection{Definitions}

We next define the notation that we will be using when describing our approach.
We denote the encoder and the decoder for the $i$th language in the system as $e_i$ and $d_i$, respectively.
For language-specific scenarios, both the encoder and decoder are considered independent modules that can be freely interchanged to work in all translation directions. 

\subsection{Language-Specific Proposed Procedure} \label{subsec:basic}

In what follows, we describe the proposed training procedure in two steps: joint training and adding new languages.

\paragraph{Joint Training} The straightforward approach is to train independent encoders and decoders for each language. The main difference from standard pairwise training is that, in this case, there is only one encoder and one decoder for each language, which will be used for all translation directions involving that language. The training algorithm for this language-specific procedure is described in Algorithm \ref{alg:train}.

\begin{algorithm}
\small
\caption{Multilingual training step}\label{alg:train}
\begin{algorithmic}[1]
\Procedure{MultilingualTrainingStep}{}
\State $N \gets \text{Number of languages in the system}$
\State $S = \{s_{0,0},...,s_{N,N}\} \gets \text{}  \{(e_i,d_j)\}$
\State $E = \{e_{0},...,e_{N}\} \gets \text{Language-specific encs.}$
\State $D = \{d_{0},...,d_{N}\} \gets \text{Language-specific decs.}$
\For{$i \gets 0$ to $N$}                    
    \For{$j \gets 0$ to $N$}                    
        \If {$s_{i,j} \in S$}
            \State $l_i, l_j = get\_parallel\_batch(i,j)$
            \State $train(s_{i,j}(e_i,d_j),l_i,l_j)$
        \EndIf    
    \EndFor
\EndFor

\EndProcedure

\medskip
\end{algorithmic}
\end{algorithm}

For each translation direction $s_{i,j}$ in the training schedule $S$ with language $i$ as source and language $j$ as target, the system is trained using the language-specific encoder $e_i$ and decoder $d_j$.  
 
\paragraph{Adding New Languages} 
Since parameters are not shared between the independent encoders and decoders, the joint training enables the addition of new languages without the need to retrain the existing modules.
Let us say we want to add language $N+1$. To do so, we must have parallel data between $N+1$ and any language in the system. For illustration, let us assume that we have $L_{N+1}-L_i$ parallel data. 
Then, we can set up a new bilingual system with language $L_{N+1}$ as source and language $L_i$ as target. To ensure that the representation produced by this new pair is compatible with the previously jointly trained system, we use the previous $L_i$ decoder ($d_{li}$) as the decoder of the new $L_{N+1}$-$L_{i}$ system and we freeze it. During training, we optimize the cross-entropy between the generated tokens and  $L_i$ reference data but update only the parameters of to the  $L_{N+1}$ encoder ($e_{l_{N+1}}$). By doing so, we train $e_{l_{N+1}}$ not only to produce good quality translations but also to produce similar representations to the already trained languages. Following the same principles, the $L_{N+1}$ decoder can also be trained as a bilingual system by freezing the $L_i$ encoder and training the decoder of the  $L_i-L_{N+1}$ system by optimizing the cross-entropy with the $L_{N+1}$ reference data.

%% file: mtexperiments.tex
In
this section we report machine translation experiments in different settings. 
Since the main difference between the shared and the language-specific encoders-decoders lies in whether they retrain the entire system when adding new languages, we accordingly design our experiments to compare the systems under this condition. 


\subsection{Data and Implementation}
We used 2 million sentences from the \textit{EuroParl} corpus \cite{koehn2005europarl} in German, French, Spanish and English as training data, with parallel sentences among all combinations of these four languages (without being multi-parallel). For Russian-English, we used 1 million training sentences from the \textit{Yandex} corpus\footnote{\url{https://translate.yandex.ru/corpus?lang=en}}. As validation and test set, we used \textit{newstest2012} and \textit{newstest2013} from WMT\footnote{\url{http://www.statmt.org}}, which is multi-parallel across all the above languages.
All data were preprocessed using standard Moses scripts \cite{koehn2007moses}

We evaluate our approach in 3 different settings: (i) the \textit{initial} training, covering all combinations of German, French, Spanish and English; (ii) \textit{adding} new languages, tested with Russian-English in both directions; and (iii) \textit{zero-shot} translation, covering all combinations between Russian and the rest of the languages.
Additionally we compare two configurations which consists in using non-tied or tied embeddings. In the language-specific approach tied embeddings consists in using language-wise word embeddings: for one language, we use the same word embeddings. Whereas, in the case of non-tied, the encoder and the decoder of each language have different word embeddings. Tied embeddings in the shared system means that both encoder and decoder share the same word embeddings.

All experiments were done using the Transformer implementation provided by Fairseq\footnote{Release v0.6.0 available at \url{https://github.com/pytorch/fairseq}}.
We used 6 layers, each with 8 attention heads,
an embedding size of 512 dimensions,
and a vocabulary size of 32k subword tokens with 
Byte Pair Encoding \cite{sennrich-etal-2016-neural} (in total for the shared encoders/decoders and per pair for language-specific encoder-decoders). Dropout was 0.1 for the shared approach and 0.3 for language-specific encoders/decoders. Both approaches were trained with an effective batch size of 32k tokens for approximately 200k updates, using the validation loss for early stopping. We used Adam \cite{kingma2014adam} as the optimizer, with learning rate of 0.001 and 4000 warmup steps. 
All experiments were performed on an NVIDIA Titan X GPU with 12 GB of memory. 

\
\begin{table}[]
\centering
\begin{tabular}{|l|cc|cc|}
\hline
      & \multicolumn{2}{c|}{Shared} & \multicolumn{2}{c|}{LangSpec}  \\ \hline 
      & $\neg$Tied &Tied & $\neg$Tied& Tied  \\ \hline
de-en & 24,40 & \textbf{{25,04}}  & 22,04 &  24,54 \\
de-es & 24,04& 25,01 & 22,38 & \textbf{{ 25,02}} \\
de-fr & 24,78 &25,14 & 22,57 & \textbf{25,49} \\
en-de & 21,39 & 21,51& 19,44 & \textbf{{ 22,01}}\\
en-es & 28,08  & 28,19& 26,79 & \textbf{29,53}  \\
en-fr & 28,43  &28,67 & 26,94 & \textbf{{ 29,74}}  \\
es-de & 19,51  & 20,21& 17,7& \textbf{{ 20,31}} \\
es-en & 26,66  & 26,93& 24,9 & \textbf{{ 27,75}} \\
es-fr & 29,47  & 29,59& 27,31  & \textbf{{ 30,08}}\\
fr-de & 19,22  & 19,81& 16,88 & \textbf{{19,97}}\\
fr-en & 25,78  & 26,29& 23,5 & \textbf{{26,55}} \\
fr-es & 28,15 & 29,03& 26,78 & \textbf{{29,07}} \\ \hline
\end{tabular}
\caption{\label{tab:baselines} Initial training. In bold, best global results.}
\end{table}



\begin{table}[]
\centering
\begin{tabular}{|l|cc|cc|}
\hline
 & \multicolumn{2}{c|}{Shared$^{RU}$} & \multicolumn{2}{c|}{LangSpec}  \\ \hline 
 & $\neg$Tied &Tied & $\neg$Tied& Tied\\ \hline
 ru-en &   24,71& 24,62 & 25,52& \textbf{{27,54}} \\
en-ru &  19,91 & 20,03 & 21,44 & \textbf{{23,94}}\\ \hline
ru-de & 15,36 & \bf 16,52 & 12,73 &  13,77\\
ru-es & 21,38 & \bf 23,12 & 18,71 &  21,08 \\
ru-fr & 21,38 & \bf 22,04  & 18,05 &  19,85\\
de-ru & 16,23 & \bf 17,27& 14,39 & {16,99}\\
es-ru & 16,98 & \bf 18,78& 15,93 &{18,46}\\
fr-ru & 16,79 & \bf 17,83&  15,16& {17.47}\\ \hline
\end{tabular}
\caption{\label{tab:add} Adding a new language translation and Zero-shot.}
\end{table}

\subsection{Shared vs Language-specific}

Table \ref{tab:baselines} and \ref{tab:add} show comparisons between the shared and language-specific encoders-decoders.

In contrast with our proposed approach, the shared system requires retraining from scratch to add a new language. For that reason, we experiment with two variants of this system: 
one trained without Russian-English (\textit{Shared}) and another one including this pair (\textit{Shared$^{RU}$}). Note that, to make experiments comparable, we use the \textit{Shared} version when comparing to our \textit{initial} system in Table \ref{tab:baselines}, and the \textit{Shared$^{RU}$} version when \textit{adding} new languages and performing \textit{zero-shot} translation.

\paragraph{Initial Training} Table \ref{tab:baselines} shows that the language-specific encoder-decoders outperforms the shared approach in all cases. On average, our proposed approach is better than the shared approach with a difference of 0.40 BLEU points.

\paragraph{Adding New Languages} Table \ref{tab:add} shows that, when adding a new language into the system, the language-specific encoder-decoders outperform the shared architecture by 2.92 BLEU points for Russian-to-English and by 3.64 BLEU in the opposite direction.  It is also worth mentioning that the Russian data is from a different domain than the frozen English modules used for training (\textit{Yandex} corpus and \textit{EuroParl}, respectively). As such, the language specific encoder-decoders are able to outperform the shared architecture when adding a new language and a new domain by learning from the previous information in the frozen modules.
Note that additionally, retraining the shared encoder-decoder to add a new language took an entire week, whereas the incremental training with the language-specific encoder-decoders was performed in only one day.

\paragraph{Zero-shot} The shared encoder-decoder clearly outperforms the language-specific encoder-decoders by 1.39 BLEU points on average. This difference in performance suggests that, while limiting the amount of shared information during training can improve our model performance, it may also harm zero-shot translation. 




Note that employing tied embeddings has a larger impact in the language-specific architecture than in the shared one. In fact, it has been key for closing the performance gap between language-specific and shared architectures.

%% file: conclusions.tex



In this paper, we present a novel method to train language-specific encoders-decoders without sharing any parameters at all. Previous works, e.g.\cite{firat2016multi,lu2018neural}, suffered from attention mismatch which was solved by sharing intermediate layers. In our case, we believe that we do not suffer from this problem because, within an initial set of \textit{N} languages, we train $N*N-1$ systems using pair-wise corpus (without requiring multi-parallel corpus as previous works \cite{escolano-etal-2019-bilingual}). Due to our proposed joint training, once we have trained our initial system, we end up with only $N$ encoders and $N$ decoders ($2*N$).

More relevantly, our system allows to incrementally add new languages into the system without having to retrain it and without varying the translation quality of initial languages in the system. When adding a new language, the language-specific encoder-decoders outperform the shared ones by 3.28 BLEU on average and, most importantly, the training of this new language was done in only one day, as opposed to the week taken by the shared system. 

%% file: acl2020.bbl
\begin{thebibliography}{15}
\expandafter\ifx\csname natexlab\endcsname\relax\def\natexlab#1{#1}\fi

\bibitem[{Arivazhagan et~al.(2019)Arivazhagan, Bapna, Firat, Lepikhin, Johnson,
  Krikun, Chen, Cao, Foster, Cherry, Macherey, Chen, and
  Wu}]{DBLP:journals/corr/abs-1907-05019}
Naveen Arivazhagan, Ankur Bapna, Orhan Firat, Dmitry Lepikhin, Melvin Johnson,
  Maxim Krikun, Mia~Xu Chen, Yuan Cao, George Foster, Colin Cherry, Wolfgang
  Macherey, Zhifeng Chen, and Yonghui Wu. 2019.
\newblock Massively multilingual neural machine translation in the wild:
  Findings and challenges.
\newblock \emph{CoRR}, abs/1907.05019.

\bibitem[{Bahdanau et~al.(2014)Bahdanau, Cho, and Bengio}]{bahdanau2014neural}
Dzmitry Bahdanau, Kyunghyun Cho, and Yoshua Bengio. 2014.
\newblock Neural machine translation by jointly learning to align and
  translate.
\newblock \emph{arXiv preprint arXiv:1409.0473}.

\bibitem[{Dong et~al.(2015)Dong, Wu, He, Yu, and Wang}]{dong-etal-2015-multi}
Daxiang Dong, Hua Wu, Wei He, Dianhai Yu, and Haifeng Wang. 2015.
\newblock Multi-task learning for multiple language translation.
\newblock In \emph{Proceedings of the ACL-IJCNLP}, pages 1723--1732, Beijing.

\bibitem[{Escolano et~al.(2019)Escolano, Costa-juss{\`a}, and
  Fonollosa}]{escolano-etal-2019-bilingual}
Carlos Escolano, Marta~R. Costa-juss{\`a}, and Jos{\'e} A.~R. Fonollosa. 2019.
\newblock From bilingual to multilingual neural machine translation by
  incremental training.
\newblock In \emph{Proceedings of the ACL: Student Research Workshop},
  Florence, Italy.

\bibitem[{Firat et~al.(2016{\natexlab{a}})Firat, Cho, and
  Bengio}]{firat2016multi}
Orhan Firat, Kyunghyun Cho, and Yoshua Bengio. 2016{\natexlab{a}}.
\newblock Multi-way, multilingual neural machine translation with a shared
  attention mechanism.
\newblock In \emph{Proceedings of the 2016 Conference of the NAACL-HLT}, pages
  866--875, San Diego, California.

\bibitem[{Firat et~al.(2016{\natexlab{b}})Firat, Sankaran, Al-Onaizan,
  Yarman~Vural, and Cho}]{firat2016zero}
Orhan Firat, Baskaran Sankaran, Yaser Al-Onaizan, Fatos~T. Yarman~Vural, and
  Kyunghyun Cho. 2016{\natexlab{b}}.
\newblock Zero-resource translation with multi-lingual neural machine
  translation.
\newblock In \emph{Proceedings of the 2016 Conference on Empirical Methods in
  Natural Language Processing}, pages 268--277, Austin, Texas.

\bibitem[{Johnson et~al.(2017)Johnson, Schuster, Le, Krikun, Wu, Chen, Thorat,
  Vi{\'e}gas, Wattenberg, Corrado et~al.}]{johnson2017google}
Melvin Johnson, Mike Schuster, Quoc~V Le, Maxim Krikun, Yonghui Wu, Zhifeng
  Chen, Nikhil Thorat, Fernanda Vi{\'e}gas, Martin Wattenberg, Greg Corrado,
  et~al. 2017.
\newblock Google’s multilingual neural machine translation system: Enabling
  zero-shot translation.
\newblock \emph{Transactions of the Association for Computational Linguistics},
  5:339--351.

\bibitem[{Kingma and Ba(2014)}]{kingma2014adam}
Diederik~P Kingma and Jimmy Ba. 2014.
\newblock Adam: A method for stochastic optimization.
\newblock \emph{arXiv preprint arXiv:1412.6980}.

\bibitem[{Koehn(2005)}]{koehn2005europarl}
Philipp Koehn. 2005.
\newblock Europarl: A parallel corpus for statistical machine translation.
\newblock In \emph{MT summit}, volume~5, pages 79--86. Citeseer.

\bibitem[{Koehn et~al.(2007)Koehn, Hoang, Birch, Callison-Burch, Federico,
  Bertoldi, Cowan, Shen, Moran, Zens et~al.}]{koehn2007moses}
Philipp Koehn, Hieu Hoang, Alexandra Birch, Chris Callison-Burch, Marcello
  Federico, Nicola Bertoldi, Brooke Cowan, Wade Shen, Christine Moran, Richard
  Zens, et~al. 2007.
\newblock Moses: Open source toolkit for statistical machine translation.
\newblock In \emph{Proceedings of the ACL: Demo Papers}, pages 177--180.

\bibitem[{Lakew et~al.(2018)Lakew, Erofeeva, Negri, Federico, and
  Turchi}]{DBLP:journals/corr/abs-1811-01137}
Surafel~Melaku Lakew, Aliia Erofeeva, Matteo Negri, Marcello Federico, and
  Marco Turchi. 2018.
\newblock Transfer learning in multilingual neural machine translation with
  dynamic vocabulary.
\newblock \emph{CoRR}, abs/1811.01137.

\bibitem[{Lu et~al.(2018)Lu, Keung, Ladhak, Bhardwaj, Zhang, and
  Sun}]{lu2018neural}
Yichao Lu, Phillip Keung, Faisal Ladhak, Vikas Bhardwaj, Shaonan Zhang, and
  Jason Sun. 2018.
\newblock A neural interlingua for multilingual machine translation.
\newblock In \emph{Proceedings of the Third Conference on Machine Translation:
  Research Papers}, pages 84--92, Belgium, Brussels.

\bibitem[{Sennrich et~al.(2016)Sennrich, Haddow, and
  Birch}]{sennrich-etal-2016-neural}
Rico Sennrich, Barry Haddow, and Alexandra Birch. 2016.
\newblock Neural machine translation of rare words with subword units.
\newblock In \emph{Proceedings of the 54th Annual Meeting of the Association
  for Computational Linguistics (Volume 1: Long Papers)}, pages 1715--1725,
  Berlin, Germany.

\bibitem[{V{\'a}zquez et~al.(2019)V{\'a}zquez, Raganato, Tiedemann, and
  Creutz}]{vazquez-etal-2019-multilingual}
Ra{\'u}l V{\'a}zquez, Alessandro Raganato, J{\"o}rg Tiedemann, and Mathias
  Creutz. 2019.
\newblock Multilingual {NMT} with a language-independent attention bridge.
\newblock In \emph{Proceedings of the RepL4NLP Workshop)}, pages 33--39,
  Florence, Italy.

\bibitem[{Zoph and Knight(2016)}]{zoph-knight-2016-multi}
Barret Zoph and Kevin Knight. 2016.
\newblock Multi-source neural translation.
\newblock In \emph{Proceedings of the 2016 Conference of the NAACL-HLT}, pages
  30--34, San Diego.

\end{thebibliography}
